\documentclass[12pt,twoside]{article}
\usepackage{amsmath,epsfig,subfigure,url}
\usepackage{algorithm, algorithmic}
\def\beq{\begin{equation}}    \def\eeq{\end{equation}}
\def\beqn{\begin{displaymath}}\def\eeqn{\end{displaymath}}
\def\bqa{\begin{eqnarray}}    \def\eqa{\end{eqnarray}}
\def\bqan{\begin{eqnarray*}}  \def\eqan{\end{eqnarray*}}
\newenvironment{keywords}{\centerline{\bf\small
Keywords}\begin{quote}\small}{\par\end{quote}\vskip 1ex}
\newtheorem{theorem}{Theorem}
\newtheorem{definition}[theorem]{Definition}
\newdimen\paravsp  \paravsp=1.3ex
\def\paradot#1{\vspace{\paravsp plus 0.5\paravsp minus 0.5\paravsp}\noindent{\bf\boldmath{#1.}}}
\def\nq{\hspace{-1em}}
\def\qed{\hspace*{\fill}\rule{1.4ex}{1.4ex}$\quad$\\}
\def\qmbox#1{{\quad\mbox{#1}\quad}}
\def\SetR{I\!\!R}
\def\fr#1#2{{\textstyle{#1\over#2}}}
\def\e{{\rm e}}                        
\def\E{{\rm I\!E}}                         
\def\P{{\rm P}}                         
\def\a{\alpha}
\def\g{\gamma}
\def\gtrsim{\buildrel{\lower.7ex\hbox{$>$}}\over{\lower.7ex\hbox{$\sim$}}}
\def\lesssim{\buildrel{\lower.7ex\hbox{$<$}}\over{\lower.7ex\hbox{$\sim$}}}

\def\D{{\cal D}}
\def\N{{\cal N}}
\def\X{{\cal X}}
\def\dist{\text{dist}}

\begin{document}
\title{\vspace{-3ex}\normalsize\sc\bf\Large\hrule height5pt \vskip 6mm
A New Local Distance-Based Outlier Detection Approach for Scattered Real-World Data
\vskip 6mm \hrule height2pt}
\author{{\bf Ke Zhang$^1$ and Marcus Hutter$^{1,2}$ and Huidong Jin$^{1,2,3}$} \\[1ex]
\normalsize $^1$RSISE, Australian National University, \\[-0.5ex]
\normalsize $^2$National ICT Australia (NICTA), Canberra Lab, ACT, Australia, \\[-0.5ex]
\normalsize $^3$CSIRO Mathematical and Information Sciences, Acton ACT 2601, Australia. \\
\normalsize \texttt{\{ke.zhang , marcus.hutter\}@rsise.anu.edu.au,\ \ Warren.Jin@csiro.au}
}
\date{March 2009}
\maketitle
\vspace{-5ex}
\begin{abstract}
Detecting outliers which are grossly different from or inconsistent
with the remaining dataset is a major challenge in real-world KDD
applications. Existing outlier detection methods are ineffective on
scattered real-world datasets due to implicit data patterns and
parameter setting issues. We define a novel \textit{Local
Distance-based Outlier Factor} (LDOF) to measure the {outlier-ness}
of objects in scattered datasets which addresses these issues. LDOF
uses the relative location of an object to its neighbours to
determine the degree to which the object deviates from its
neighbourhood.
Properties of LDOF are theoretically analysed including LDOF's lower
bound and its false-detection probability, as well as parameter
settings. In order to facilitate parameter settings in real-world
applications, we employ a top-$n$ technique in our outlier detection
approach, where only the objects with the highest LDOF values are
regarded as outliers. Compared to conventional approaches (such as
top-$n$ KNN and top-$n$ LOF), our method top-$n$ LDOF is more
effective at detecting outliers in scattered data. It is also easier
to set parameters, since its performance is relatively stable over a
large range of parameter values, as illustrated by experimental
results on both real-world and synthetic datasets.
\def\contentsname{\centering\normalsize Contents}
{\parskip=-2.5ex\tableofcontents}
\end{abstract}

\vspace{-2ex}

\section{Introduction}\label{sec:introduction}

Of all the data mining techniques that are in vogue, outlier
detection comes closest to the metaphor of mining for nuggets of
information in real-world data. It is concerned with
discovering the exceptional behavior of certain
objects~\cite{Tang:02}. Outlier detection
techniques have widely been applied in medicine (e.g. adverse
reactions analysis), finance (e.g. financial fraud detection),
security (e.g. counter-terrorism), information security (e.g.
intrusions detection) and so on. In the recent decades, many outlier
detection approaches have been proposed, which can be broadly
classified into several categories:
distribution-based~\cite{Barnett:94}, depth-based~\cite{Tukey:77},
distance-based (e.g. KNN)~\cite{Knorr:98},
cluster-based (e.g. DBSCAN)~\cite{Ester:96} and
density-based (e.g. LOF)~\cite{Breunig:00}
methods.

However, these methods are often unsuitable in real-world
applications due to a number of reasons. Firstly, real-world data
usually have a scattered distribution, where objects are loosely
distributed in the domain feature space. That is, from a `local'
point of view, these objects cannot represent explicit patterns
(e.g. clusters) to indicate normal data `behavior'. However, from a
`global' point of view, scattered objects constitute several
{mini-clusters}, which represent the pattern of a subset of objects.
Only the objects which do not belong to any other object groups are
genuine outliers. Unfortunately, existing outlier definitions depend
on the assumption that most objects are crowded in a few main
clusters. They are incapable of dealing with scattered datasets,
because {mini-clusters} in the dataset evoke a high
false-detection rate (or low precision).

Secondly, it is difficult in current outlier detection approaches to
set accurate parameters for real-world datasets . Most
outlier algorithms must be tuned through
{trial-and-error}~\cite{Fan:06}. This is
impractical, because real-world data usually do not contain
labels for anomalous objects. In addition, it is hard to evaluate
detection performance without the confirmation of domain experts.
Therefore, the detection result will be uncontrollable if parameters
are not properly chosen.

To alleviate the parameter setting problem, researchers proposed
top-$n$ style outlier detection methods. Instead of a binary outlier
indicator, top-$n$ outlier methods provide a ranked list of objects
to represent the degree of {`outlier-ness'} for each object. The
users (domain experts) can {re-examine} the selected top-$n$ (where
$n$ is typically far smaller than the cardinality of dataset)
anomalous objects to locate real outliers. Since this detection
procedure can provide a good interaction between data mining experts
and users, top-$n$ outlier detection methods become popular in
real-world applications.

Distance-based, top-$n$ {$K^{th}$-Nearest} Neighbour
distance~\cite{Ramaswamy:00} is a typical top-$n$ style outlier
detection approach. In order to distinguish from the original
{distance-based} outlier detection method in~\cite{Knorr:98}, we
denote {$K^{th}$-Nearest} Neighbour distance outlier as {top-$n$
KNN} in this paper. In {top-$n$ KNN} outlier, the distance from an
object to its $k^{th}$ nearest neighbour (denoted as {$k$-distance}
for short) indicates {outlier-ness} of the object. Intuitively, the
larger the {$k$-distance} is, the higher {outlier-ness} the object
has. {Top-$n$ KNN} outlier regards the $n$ objects with the highest
values of {$k$-distance} as outliers~\cite{Ramaswamy:00}.

A {density-based} outlier, Local Outlier Factor
(LOF)~\cite{Breunig:00}, was proposed in the same
year as top-$n$ KNN. In LOF, an outlier factor is assigned
for each object w.r.t its surrounding neighbourhood. The outlier
factor depends on how the data object is closely packed in its
locally reachable neighbourhood~\cite{Fan:06}.
Since LOF uses a threshold to differentiate outliers from normal
objects~\cite{Breunig:00}, the same problem of parameter setting
arises. A lower {outlier-ness} threshold will produce high
false-detection rate, while a high threshold value will
result in missing genuine outliers.
In recent real-world applications, researchers have found it
more reliable to use LOF in a top-$n$ manner~\cite{Tang:02},
i.e.\ only objects with the highest LOF values will be considered
outliers. Hereafter, we call it top-$n$ LOF.

Besides top-$n$ KNN and top-$n$ LOF, researchers have proposed other
methods to deal with real-world data, such as the
{connectivity-based} (COF)~\cite{Tang:02}, and Resolution
{cluster-based} ({RB-outlier})~\cite{Fan:06}. Although the existing
top-$n$ style outlier detection techniques alleviate the difficulty
of parameter setting, the detection precision of these methods (in
this paper, we take {top-$n$ KNN} and top-$n$ LOF as typical
examples) is low on scattered data. In Section~\ref{sec:problem}, we
will discuss further problems of top-$n$ KNN and top-$n$ LOF.

In this paper we propose a new outlier detection definition, named
Local Distance-based Outlier Factor (LDOF), which is sensitive to
outliers in scattered datasets. LDOF uses the relative distance from
an object to its neighbours to measure how much objects deviate from
their scattered neighbourhood. The higher the violation degree an
object has, the more likely the object is an outlier. In addition,
we theoretically analyse the properties of LDOF, including its lower
bound and false-detection probability, and provide guidelines for
choosing a suitable neighbourhood size. In order to simplify
parameter setting in real-world applications, the top-$n$ technique
is employed in our approach. To validate LDOF, we perform various
experiments on both synthetic and real-world datasets, and compare
our outlier detection performance with top-$n$ KNN and top-$n$ LOF.
The experimental results illustrate that our proposed top-$n$ LDOF
represents a significant improvement on outlier detection capability
for scattered datasets.

The paper is organised as follows: In Section~\ref{sec:problem}, we
illustrate and discuss the problems of top-$n$ KNN and top-$n$ LOF
on a real-world data. In Section~\ref{sec:definition}, we formally
introduce the outlier definition of our approach, and mathematically
analyse properties of our {outlier-ness} factor in
Section~\ref{sec:property}. In Section~\ref{sec:algorithm}, the
top-$n$ LDOF outlier detection algorithm is described, together with
an analysis of its complexity. Experiments are reported in
Section~\ref{sec:experiment}, which show the superiority of our
method to previous approaches, at least on the considered datasets.
Finally, conclusions are presented in Section~\ref{sec:conclusion}.

\section{Problem Formulation}\label{sec:problem}
\footnotetext[1]{WDBC dataset is from UCI ML Repository:
http://archive.ics.uci.edu/ml.}

\begin{figure*}[t]
\centering \subfigure[{2-D} projection of WDBC.]%
{\epsfig{file=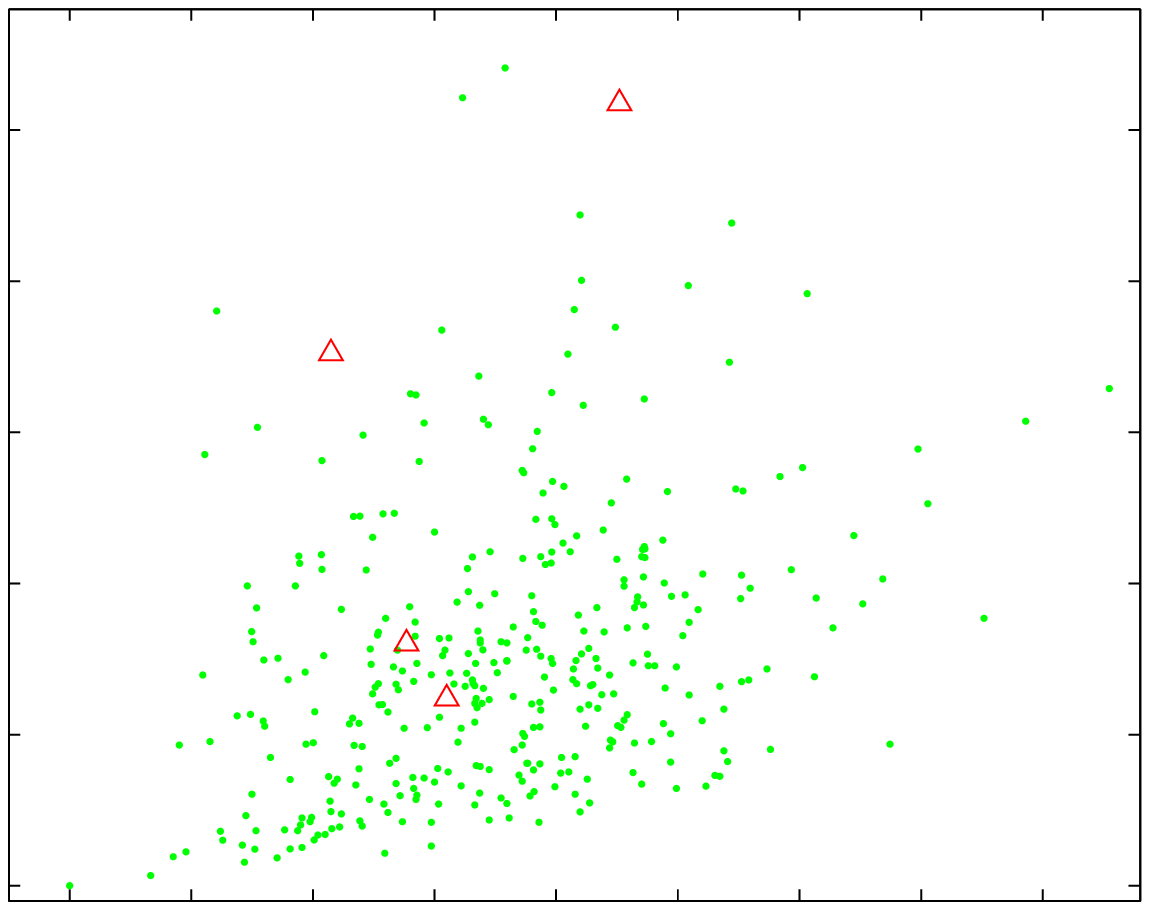,width=0.45\linewidth,height=0.33\linewidth}}\hfill
\subfigure[Synthetic {2-D} data.]%
{\epsfig{file=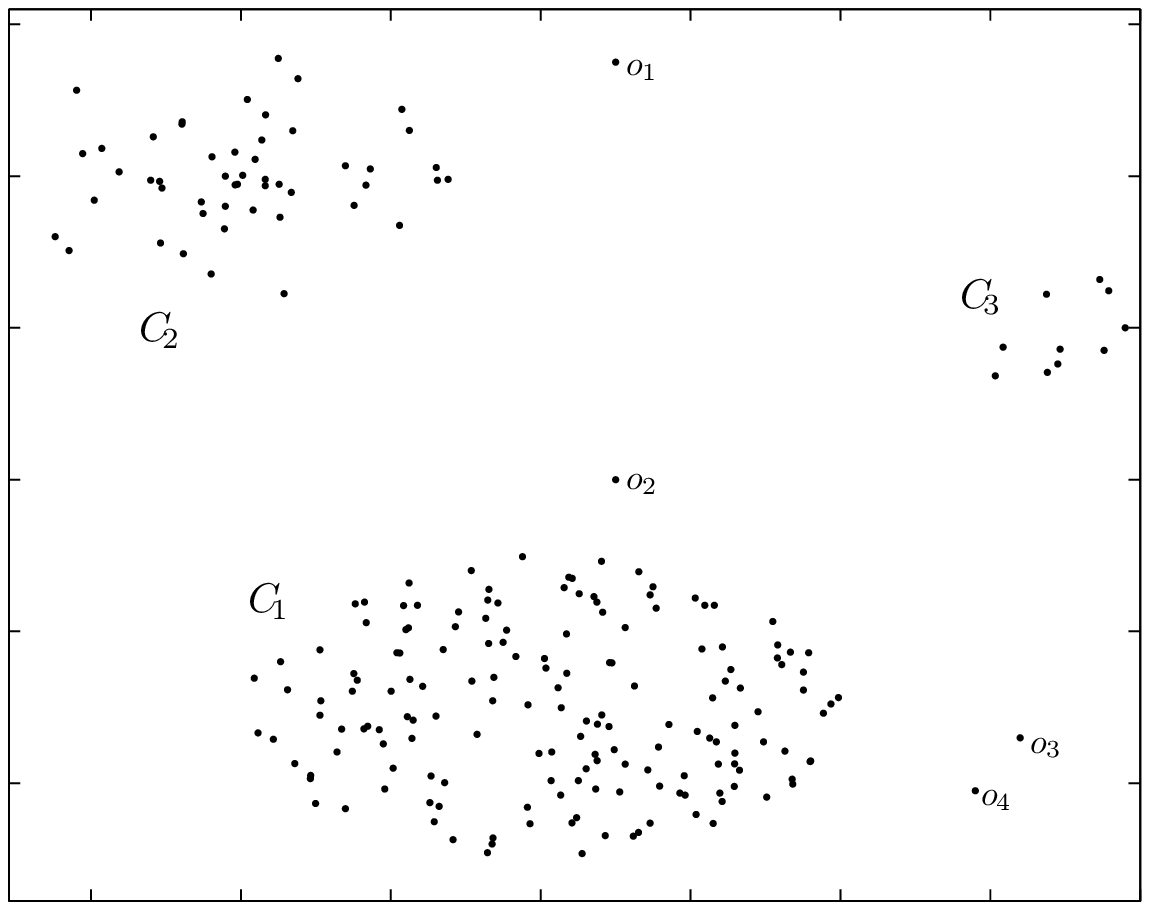,width=0.45\linewidth,height=0.33\linewidth}}
 \caption{(a) The 2-D projection of a real-world dataset.
 (b) Simple 2-D illustration.} \label{fig:Demon}
\end{figure*}

In real-world datasets, high dimensionality (e.g. 30 features) and
sparse feature value range usually cause objects to be scattered in
the feature space. The scattered data is similar to the distribution
of stars in the universe. Locally, they seem to be randomly
allocated in the night sky (i.e. stars observed from the Earth),
whereas globally the stars constitute innumerable galaxies.
Figure~\ref{fig:Demon}(a) illustrates a 2-D projection of a
real-world dataset, \textit{Wisconsin Diagnostic Breast Cancer}
(WDBC)\footnotemark[1], which is typically 30-D. The green points
are the benign diagnosis records (regarded as normal objects), and
the red triangles are malignant diagnosis records (i.e.\ outliers we
want to capture). Obviously, we cannot detect these outliers in 2-D
space, whereas in high dimension (e.g. 30-D), these scattered normal
objects constitute a certain number of loosely bounded
mini-clusters, and we are able to isolate genuine outliers. Unlike
galaxies, which always contain billions of stars, these
mini-clusters in scattered datasets usually have a relatively small
number of objects. Figure~\ref{fig:Demon}(b) is a simple
demonstration of this situation, where $C_1$ is a {well-shaped}
cluster as we usually define in other outlier detection methods.
$C_2$ and $C_3$ are comprised of scattered objects with loose
boundary, called mini-clusters. These small clusters should be
recognised as `normal', even if they contain a small number of
objects. The objects of our interest are the points lying far away
from other {mini-clusters}. Intuitively, $o_1$, $o_2$, $o_3$, $o_4$
are outliers in this sample. We recall a well accepted informal
outlier definition proposed by Hawkins~\cite{Hawkins:80}:
\textit{``An outlier is an observation that deviates so much from
other observations as to arouse suspicion that it was generated by a
different mechanism''}. In scattered datasets, an outlier should
be an object deviating from any other group of objects.

The only way in which our outlier definition differs from others
(e.g. in ~\cite{Knorr:98} and~\cite{Breunig:00}) is that the normal
pattern of data is represented by scattered objects, rather than
crowded main clusters. The neighbourhood in scattered real-world
datasets has two characteristics: (1) objects in mini-clusters are
loosely distributed; (2) when neighbourhood size $k$ is large, two
or more mini-clusters are taken into consideration. The
neighbourhood becomes sparse as more and more objects which belong
to different mini-clusters should be taken into account.

As discussed above, top-$n$ KNN and top-$n$ LOF are ineffective
for scattered datasets. Take a typical example, in
Figure~\ref{fig:Demon}(b), when $k$ is greater than the cardinality
of $C_3$ (10 in this case), some objects in $C_1$ become neighbours
of the objects in $C_3$. Hence, for top-$n$ KNN, the
{$k$-distance} of the object can be larger than genuine outliers.
For top-$n$ LOF, since the density of $C_3$ is smaller than that
of $C_1$, it also fails for ranking $o_1$, $o_2$, $o_3$ and $o_4$ in
the highest {outlier-ness} positions. In
Section~\ref{sec:experiment}, we will demonstrate that the two methods
fail to detect genuine outliers when $k$ grows greater than 10.

Intuitively, it is more reasonable to measure how an object deviates
from its neighbourhood system as an {outlier-ness} factor
rather than global distance (top-$n$ KNN) or local density
(top-$n$ LOF). Thereby, we propose LDOF to measure the degree
of neighbourhood violation. The formal definition of LDOF is
introduced in the following section.

\begin{figure*}[t]
\centering
\subfigure[]{\epsfig{file=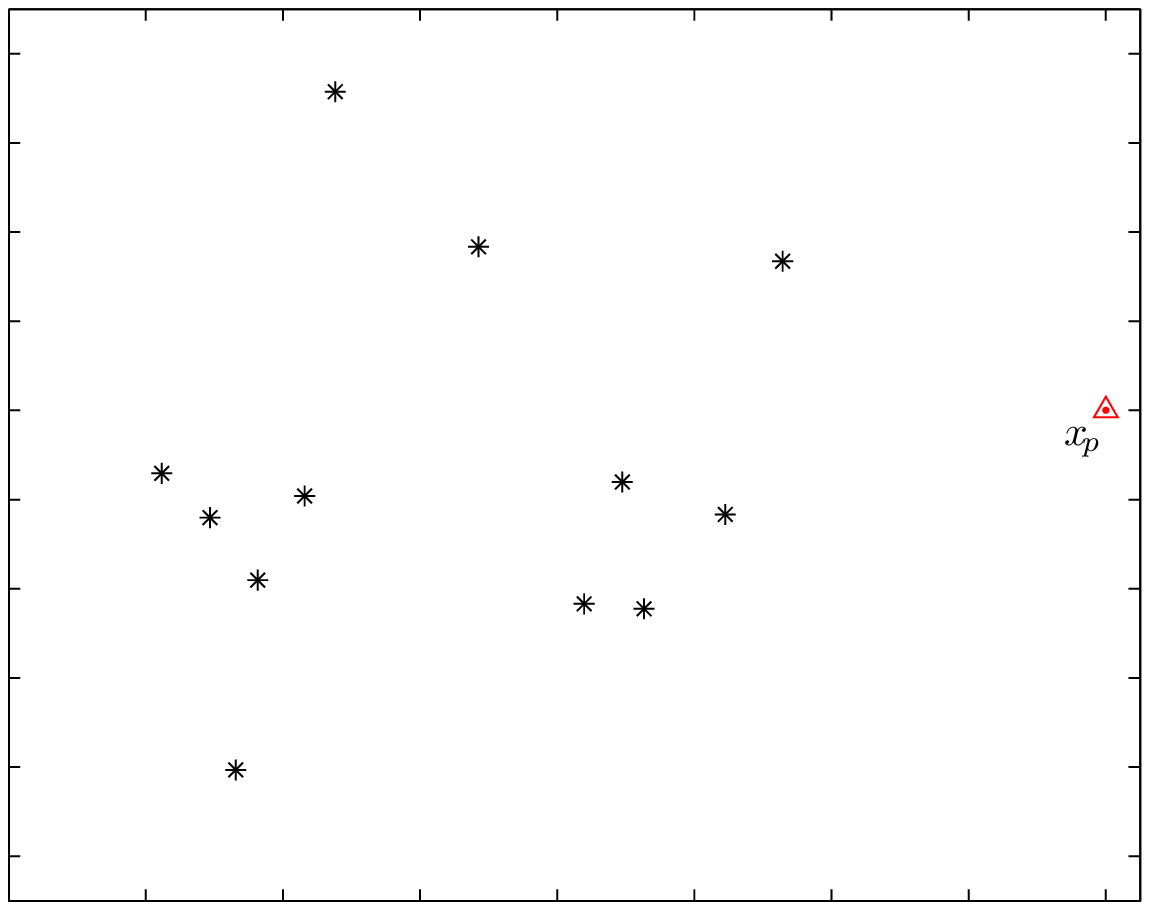,width=0.45\linewidth,height=0.37\linewidth}}\hfill
\subfigure[]{\epsfig{file=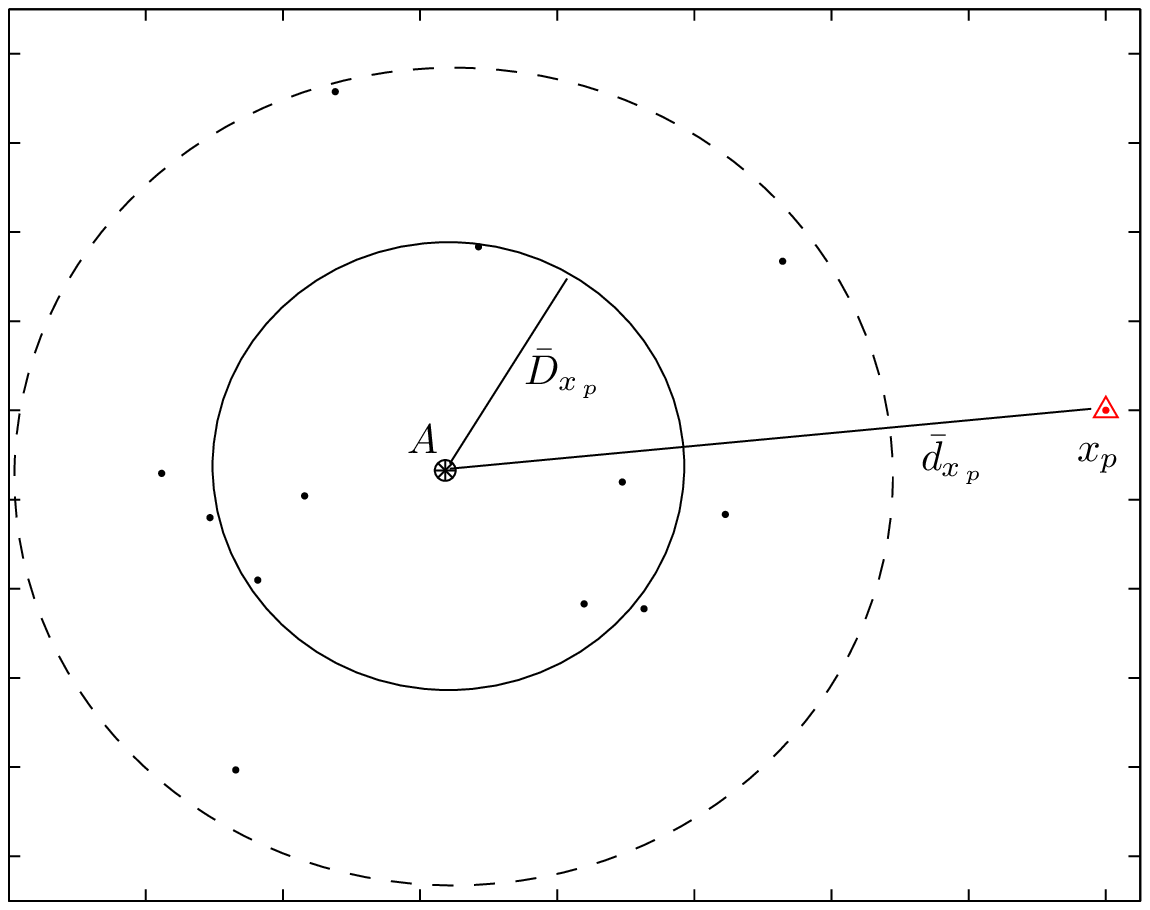,width=0.45\linewidth,height=0.37\linewidth}}
\caption{(a) An anomalous object $x_p$ with scattered neighbours.
(b) The explicit {outlier-ness} of object $x_p$ with the help of
LDOF definition. $A$ is the center of neighbourhood system of $x_p$.
The dashed circle includes all neighbours of $x_p$. The solid circle
is $x_p$'s ``reformed'' neighbourhood region.} \label{fig:neighbor}
\end{figure*}

\section{Formal Definition of Local Distance-based Outliers}\label{sec:definition}

In this section, we develop a formal definition of the Local
Distance-based Outlier Factor, which avoids the shortcomings
presented above.

\begin{definition}[KNN distance of $x_p$]\label{def:KNNdist}
Let $\N_p$ be the set of the {$k$-nearest} neighbours
of object $x_p$ (excluding $x_p$). The {$k$-nearest} neighbours
distance of $x_p$ equals the average distance from $x_p$ to all
objects in $\N_p$.
More formally, let $\dist(x,x')\geq 0$ be a distance measure between
objects $x$ and $x'$. The $k$-nearest neighbours
distance of object $x_p$ is defined as
\beqn
  \bar d_{x_p} \;:=\; \frac{1}{k} \sum_{x_i \in \N_p} \dist(x_i,x_p).
\eeqn
\end{definition}

\begin{definition}[KNN inner distance of $x_p$]\label{def:KNNinnerDist}
Given the {$k$-nearest} neighbours set $\N_p$ of
object $x_p$, the $k$-nearest neighbours inner distance of
$x_p$ is defined as the average distance among objects in
$\N_p$:
\beqn
  \bar D_{x_p} \;:=\; \frac{1}{k(k - 1)} \sum_{x_i, x_{i'} \in
  \N_p, i \neq i'} \dist(x_i, x_{i'}).
\eeqn
\end{definition}

\begin{definition}[LDOF of $x_p$]\label{def:LDOF}
The local {distance-based} outlier factor of $x_p$ is defined as:
\beqn
  LDOF_k (x_p) \;:=\; \frac{\bar d_{x_p}}{\bar D_{x_p}}
\eeqn
\end{definition}

If we regard the {$k$-nearest} neighbours as a neighbourhood
system, $LDOF$ captures the degree to which object $x_p$ deviates from
its neighbourhood system. It has the clear intuitive meaning that
$LDOF$ is the distance ratio indicating how far the object $x_p$
lies outside its neighbourhood system. When $LDOF\lesssim 1$, it means
that $x_p$ is surrounded by a data `cloud'. On the contrary, when
$LDOF \gg 1$, $x_p$ is outside the whole `cloud'. It is easy to see
that the higher $LDOF$ is, the farther $x_p$ is away from its
neighbourhood system.

To further explain our definition, we exemplify it in Euclidian
space. Hereinafter, let $x_i \in \X = \SetR^d$, and $\bar x
:= \fr1k \sum_{x_i \in \N_p} x_i$. For the squared Euclidian
distance $||\cdot||^2$, the outlier definition can be written as:
\bqa
  \bar d_{x_p} &=& \frac{1}{k} \sum_{x_i \in \N_p} ||x_p-x_i||^2
  \;=\; ||x_p -\bar x||^2 + \frac{1}{k} \sum_{x_i \in \N_p}||x_i - \bar x||^2,
\\
  \bar D_{x_p} &=& \frac{1}{k(k-1)} \sum_{x_i,x_{i'} \in \N_p, i \neq i'\nq\nq} ||x_i - x_{i'}||^2
  \;=\; \frac{2}{k -1} \sum_{x_i \in \N_p} ||x_i - \bar x||^2.
\label{eq:EuclidianDist}
\eqa
Thus, $LDOF_k(x_p)\gg 1$, i.e.\ $x_p$ lies outside its neighbourhood
system, iff
\beq
  ||x_p - \bar x||^2 \;\gg\; \frac{k + 1}{k(k - 1)} \sum_{x_i \in
  \N_p} ||x_i - \bar x||^2. \label{eq:EuclidianOutlier}
\eeq
The same expression holds for the more general Mahalanobis
distance~\cite{Mardia:79}. In Equation~\ref{eq:EuclidianOutlier},
the {lefthand-side} is the square distance of $x_p$ to its
neighbourhood centroid $\bar x$, and the {righthand-side} becomes
the distance variance in $\N_p$ when $k \gg 1$. Therefore,
Equation 6 can be understood as follows: The {$k$-nearest}
neighbours of object $x_p$ form a ``reformed'' neighbourhood
region, represented as a hyperball with radius $\bar D_{x_p}$,
centered at $\bar x$. As illustrated in
Figure~\ref{fig:neighbor}(a), since the neighbours of $x_p$ are
scattered, it is unclear whether $x_p$ (indicated by
$\bigtriangleup$) belongs to its neighbourhood system or not.
Our $LDOF$ definition, as shown in Figure~\ref{fig:neighbor}(b), it
clearly regards $x_p$ as lying outside its reformed neighbourhood
region. The $LDOF$ of $x_p$ is obviously greater than 1, which
indicates that $x_p$ is an outlier. Through this example we can see
that $LDOF$ can effectively capture the {outlier-ness} of an object
among a scattered neighbourhood. In addition, as $k$ grows, $LDOF$
takes more objects into consideration, and the view of $LDOF$
becomes increasingly global. If an object is far from its large
neighbourhood system (extremely the whole dataset) it is definitely
a genuine outlier. Hence, the detection precision of our method
might be stable over a large range of $k$. In the following section,
we will theoretically analyse properties of $LDOF$, and propose a
heuristic for selecting the neighbourhood size $k$.

\section{Properties of LDOF}\label{sec:property}

\paradot{Lower bound of $LDOF$}
Ideally, we prefer a universal threshold of $LDOF$ to unambiguously
distinguish abnormal from normal objects (e.g. in any datasets, an
object is outlier if $LDOF>1$). However, the threshold is problem
dependent due to the complex structure of real-world datasets. Under
some continuity assumption, we can calculate an asymptotic lower
bound on $LDOF$, denoted as $LDOF_{lb}$. $LDOF_{lb}$ indicates that
an object is an inlier (or normal) if its $LDOF$ is smaller than
$LDOF_{lb}$.

\begin{theorem}[$LDOF$ {lower-bound} of outliers]\label{thmCUD}
Let data $\D$ be sampled from a density that is
continuous at $x_p$. For $N\gg k\gg 1$ we have
$LDOF_{lb}\approx\fr12$ with high probability.
More formally, for $k,N\to\infty$ such that the neighbourhood
size $\bar D_{x_p}\to 0$ we have
\beqn
  LDOF_{lb} = \frac{\bar d_{x_p}}{\bar D_{x_p}} \to \frac{1}{2}
 \qmbox{with probability 1}
\eeqn
\end{theorem}
The theorem shows that when $LDOF\approx\fr12$, the point is
squarely lying in a uniform cloud of objects, i.e. it is not an
outlier. The lower-bound of $LDOF$ provides a potential pruning rule
of algorithm complexity. In practice, objects can be directly
ignored if their $LDOF$s are smaller than $\fr12$. Remarkably,
$LDOF_{lb}$ does not depend on the dimension of $\X$. This is very
convenient: data often lie on lower-dimensional manifolds. Since
locally, a manifold is close to an Euclidian space (of lower
dimension), the result still holds in this case. Therefore, we do
not need to know the effective dimension of our data.

\paradot{Proof sketch}
Consider data sampled from a continuous density (e.g. Gaussian or
other standard distributions). For fixed $k$, as sample size $N$
goes to infinity, the size of the $k$-nearest neighbours region
tends to zero. Locally any continuous distribution is approximately
uniform. In the following we assume a uniform density around $x_p$.
The achieved result then generalizes to arbitrary distributions
continuous at $x_p$ by taking the limit $N\to\infty$.

Without loss of generality, let $x_p=0$. Fix some sufficiently small
radius $r>0$ and let $B_r$ be the ball of radius $r$ around $0$. By
assumption, data $\D$ is locally uniformly distributed,
which induces a uniform distribution in $B_r$, i.e.\ all $x_i\in
\N_p$ are uniformly distributed random variables in $B_r$.
Hence their expected value $\E[x_i]=0$. This implies
\bqa\label{eq:T1expectation}
  \E[\bar D_{x_p}] &=& \frac{1}{k(k-1)} E \Big[\sum_{i \neq i'}(||x_i||^2-2x_i \cdot x_{i'}+||x_{i'}||^2)\Big]
\\ \nonumber
  &=& \frac{2}{k}\sum_{x_i\in \N_p} E [||x_j||^2]
  \;=\; 2 \E[\bar d_{x_p}].
\eqa
In the first equality we simply expanded the square in the
definition of $\bar D_{x_p}$, where $\cdot$ is the scalar product.
In the second equality we used $\E[x_i \cdot x_{i'}]=\E[x_i] \cdot
\E[x_{i'}] = 0$ for $i \neq i'$. The last equality is just the
definition of $\bar d_{x_p}$ for $x_p = 0$. Taking the ratio we get
\beqn
  \E[\bar d_{x_p}] / \E[\bar D_{x_p}] = 1 / 2.
\eeqn
Note that the only property of the sampling distribution we used was
$\E[x_i]=0$, i.e. the result holds for more general distributions
(e.g.\ any symmetric distribution around $x_p=0$).

Using the central limit theorem or explicit calculation, one can
show that for large $k$ and $N$, the distributions of $\bar d_{x_p}$ and
$\bar D_{x_p}$ concentrate around their means $\E[\bar d_{x_p}]$ and
$\E[\bar D_{x_p}]$, respectively, which implies that $\bar d_{x_p} /
\bar D_{x_p} \approx 1/2$ with high probability.

This also shows that for any sampling density continuous at $x_p$
(since they are locally approximately uniform), $\bar d_{x_p} / \bar
D_{x_p} \to \fr12$ holds, provided $\bar D_{x_p}\to 0$. We
skip the formal proof. \qed

\paradot{False-detection probability}
As discussed in Section~\ref{sec:introduction}, in real-world
datasets, it is hard to set parameters properly by trial-and-error.
Instead of requiring prior knowledge from datasets (e.g. outlier
labels), we theoretically determine the false-detection probability,
given neighbourhood size $k$.

\begin{theorem}[{False-detection} probability of LDOF]\label{th:ErrorProbability}
Let data $\D$ be uniformly distributed in a neighbourhood of $x_p$
containing $k$ objects $\N_p$. For $LDOF$ threshold $c>\fr12$, the
probability of false detecting $x_p\in\SetR^d$ as an outlier is
exponentially small in $k$. More precisely,
\beqn 
  \P[LDOF_k(x_p)>c] \;<\; \e^{-\a (k-2)},\qmbox{where}
  \a:=\textstyle {2\over 25}(1-{1\over 2c})^2({d\over d+2})^2
\eeqn
The bound still holds for non-uniform densities continuous in $x_p$,
provided $N\gg k$.
\end{theorem}
In particular, for $c=1$ in high-dimensional spaces ($d\to\infty$)
we get $\a\to\fr{1}{50}$. So for $k\gg 50$ the false-detection
probability is very small. Note that because the bound is quite
crude, we can expect good performance in practice for much smaller
$k$. On the other hand, choosing $c\approx\fr12$ degenerates the
bound (i.e. $\a \to 0$), consistent with Theorem~\ref{thmCUD}.

\paradot{Proof sketch}
We follow the notation used in the proof of Theorem~\ref{thmCUD}.
We consider a uniform data distribution first.
For $x_p=0$ and dropping $p$, we can write the distances as
\beqn
  \bar d = \overline{x^2},\qquad
  \bar D = {2k\over k-1}(\overline{x^2}-\bar x^2),\qquad
  \bar x^2:=||{1\over k}\sum_{j\in\N}x_j||^2,\qquad
  \overline{x^2} := {1\over k}\sum_{j\in\N}||x_j||^2
\eeqn
For $x_j$ uniformly distributed in ball $B_r:=\{x:||x||\leq r\}$,
one can compute the mean square length explicitly:
\beq\label{eqa}
  a:=\E[||x_j||^2] \;=\; {\int_{B_r}||x||^2\; dx\over\text{Volume}{(B_r)}}
  \;=\; {\int_0^r r^2 r^{d-1} dr\over \int_0^r r^{d-1}dr}
  \;=\; {d\over d+2}\, r^2
\eeq
where $d$ is the dimensionality of $x=x_j\in\X=\SetR^d$. The first
equality is just the definition of a uniform expectation over $B_r$.
The second equality exploits rotational symmetry and reduces the
$d$-dimensional integral to a one-dimensional radial integral. The
last equality is elementary. The expected values of
$\overline{x^2}$ and $\bar x^2$, respectively, are
\bqa\label{eqa2}
  \E[\overline{x^2}] &=& {1\over k}\sum_j\E[||x_j||^2] \;=\; a
\\ \label{eqak}
  \E[\bar x^2] &=& {1\over k^2}\sum_{j,j'}\E[x_j\cdot x_{j'}]
  \;=\; {1\over k^2}\sum_j\E[x_j^2] \;=\; {1\over k}a
\eqa
where we have exploited $\E[x_j\cdot
x_{j'}]=\E[x_j]\cdot\E[x_{j'}]=0$ for $j\neq j'$.
By rearranging terms, we see that
\beqn
  \bar d>c\bar D \quad\Leftrightarrow\quad \bar x^2 >\g\overline{x^2},
  \qmbox{where} \fr1k < \g:=1-{k-1\over 2kc} < 1\quad (c>\fr12)
\eeqn
Thus we need (bounds on) the probabilities that $\overline{x^2}$ and
$\bar x^2$ deviate (significantly) from their expectation. For any
(vector-valued) i.i.d.\ random variables $x_1,...,x_k$ and
any function $f(x_1,...,x_k)$ symmetric under permutation of its
arguments, McDiarmid's inequality
can be written as follows:
\bqan
  & &\mbox{Let}\quad \Delta \;\geq\; \Delta':=\sup_{x_2..x_k}\{\sup_{x_1} f(x_1,...,x_k)-\inf_{x_1}f(x_1,...,x_k)\},\quad\mbox{then}
\\
  & & \P[f(x_1,...,x_k)-\E[f(x_1,...,x_k)]\geq t]
  \;\leq\; \exp\{-2t^2/k\Delta\} \quad\forall t\geq 0
\eqan
For $f_1:=\bar x^2$ an elementary calculation using $x_j\in B_r$
gives $\Delta'_1=4(k-1)r^2/k^2$. For $f_2:=\overline{x^2}$ we get
$\Delta'_2=r^2/k$ straightforwardly.
Now consider the real quantity of interest: $f(x_1,...,x_k):=\bar
x^2-\g\overline{x^2}$. Combining the ranges, we can bound
$\Delta'\leq\Delta'_1+\g\Delta'_2\leq 5r^2/k=:\Delta$. The
expectation of $f$ is $\E[\bar x^2-\g\overline{x^2}]=\fr1k a-\g a$.
Let $t:=a(\g-\fr1k)>0$. Then using McDiarmid's inequality we get
\bqan
  \P[\bar d>c\bar D]
  &=& \P[\bar x^2 >\g\overline{x^2}]
  \;=\; \P[(\bar x^2-\g\overline{x^2}) -\E[\bar x^2-\g\overline{x^2}] \geq t]
\\
  &\leq& \exp\{-2t^2/k\Delta^2\} \;\leq \exp\{-\a (k-2)\}
\eqan
The last inequality follows from
\beqn
  {2t^2\over k\Delta^2}
  \;=\; {2a^2k\over 25r^4}\Big(\g-{1\over k}\Big)^2
  \;=\; {2k\over 25}\Big({d\over d+2}\Big)^2
        \left[\Big(1-{1\over k}\Big)\Big(1-{1\over 2c}\Big)\right]^2
  \;\geq\; \a
\eeqn
where we have inserted $\Delta$, $a$, and $\g$, and used
$k(1-\fr1k)^2\geq k-2$ and $\a$ from the theorem.
This proves the theorem for uniform distribution.

An analogous argument as in the proof of Theorem~\ref{thmCUD} shows
that the result still holds for non-uniform distributions if
$N\to\infty$, since a continuous density is locally
approximately uniform. \qed

\section{LDOF Outlier Detection Algorithm and Its Complexity}\label{sec:algorithm}

\paradot{{Top-$n$} LDOF}
Even with the theoretical analysis of the previous section, it is
still hard to determine a threshold for $LDOF$ to identify outliers
in an arbitrary dataset. Therefore we employ top-$n$ style
outlier detection, which ranks the $n$ objects with the highest
$LDOF$s. The algorithm that obtains the top-$n$ $LDOF$ outliers
for all the $N$ objects in a given dataset $\D$ is outlined in
Algorithm 1.

\paradot{How to choose $k$}
Based on Theorem~\ref{th:ErrorProbability}, it is beneficial to use
a large neighbourhood size $k$. However, too large $k$ will lead to
a global method with the same problems as top-$n$ KNN outlier. For
the best use of our algorithm, the lower bound of potentially
suitable $k$ is given as follows: If the effective dimension of the
manifold on which $\D$ lies is $m$, then at least $m$ points are
needed to `surround' another object. That is to say a $k>m$
is needed. In Section~\ref{sec:experiment}, we will see that, when
$k$ increases to the dimension of the dataset, the detection
performance of our method rises, and remains stable for a wide range
of $k$ values. Therefore, the parameter $k$ in LDOF is easier to
choose than in other outlier detection approaches.

\begin{algorithm}[t]
\caption{{Top-$n$} LDOF ({Top-$n$} Local Distance-based Outlier Factor)}\label{alg:LDOF}
{\bf Input:} A given dataset $\D$, natural numbers $n$ and $k$.
\begin{enumerate}
\item For each object $p$ in $\D$, retrieve $p$'s {$k$-nearest} neighbours;
\item Calculate the $LDOF$ for each object $p$. \\
      The objects with $LDOF < LDOF_{lb}$ are directly discarded;
\item Sort the objects according to their $LDOF$ values;
\item {\bf Output:} the first $n$ objects with the highest $LDOF$ values.
\end{enumerate}
\end{algorithm}

\paradot{Algorithm complexity}
In Step 1, querying the {$k$-nearest} neighbours, takes the majority
of the computational load. Naively, the runtime of this step is
$O(N^2)$. If a {tree-based} spatial index such as {$X$-tree} or
{$R^*$-tree} is used~\cite{Breunig:00,Breunig:99}, the complexity is
reduced to $O(N \log N)$. Step 2 is straightforward and calculates
$LDOF$ values according to Definition \ref{def:LDOF}. As the
$k$-$nn$ query is materialised, this step is linear in $N$. Step 3
sorts the $N$ objects according to their $LDOF$ values, which can be
done in $O(N \log N)$. Since the objects with $LDOF < LDOF_{lb}$ are
flushed (i.e.\ they are definitely {non-outliers}), the number of
objects needed to sort in this step is smaller than $N$ in practice.
Finally, the overall computation complexity of Algorithm 1 is $O(N
\log N)$ with appropriate index support.

\section{Experiments}\label{sec:experiment}

In this section, we compare the outlier detection performance of
top-$n$ LDOF with two typical top-$n$ outlier detection methods,
top-$n$ KNN and top-$n$ LOF. Experiments start with a synthetic
{2-D} dataset which contains outliers that are meaningful but are
difficult for top-$n$ KNN and top-$n$ LOF. In Experiments 2 and
3, we identify outliers in two real-world datasets to illustrate
the effectiveness of our method in real-world situations. For
consistency, we only use the parameter $k$ to represent the
neighbourhood size in the investigation of the three methods. In
particular, in top-$n$ LOF, the parameter {\it MinPts} is set to
neighbourhood size $k$ as chosen in the other two methods.

\footnotetext[2]{Precision$=n_{\text{real-outliers in {top-n}}}/n$.
We set $n$ as the number of real outliers if possible.}

\begin{figure*}[t]
\centering
\subfigure[Precisions in synthetical dataset.]%
{\epsfig{file=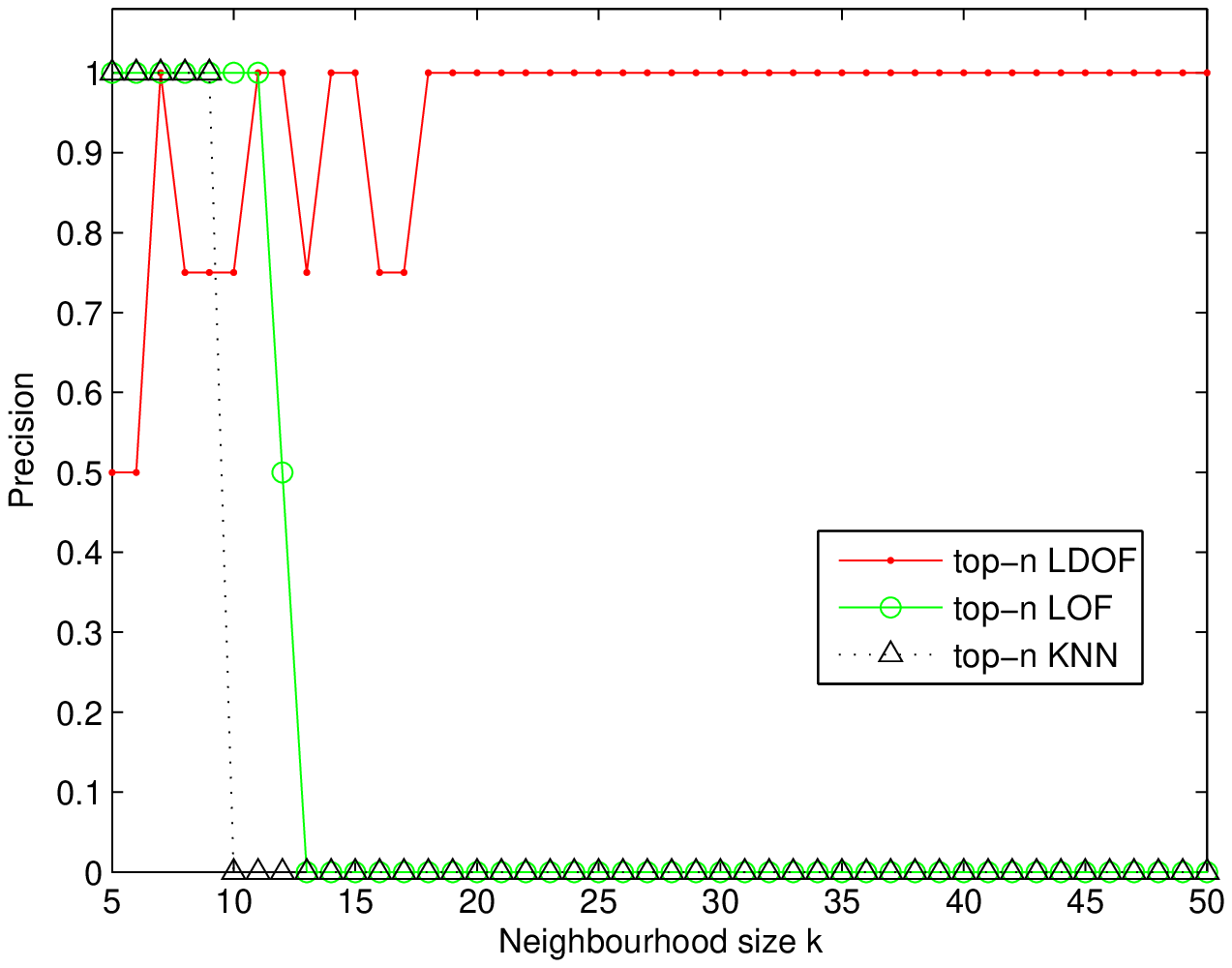,width=0.45\linewidth,height=0.35\linewidth}}\label{fig:neighborNC}\hfill
\subfigure[Precisions in WDBC dataset.]%
{\epsfig{file=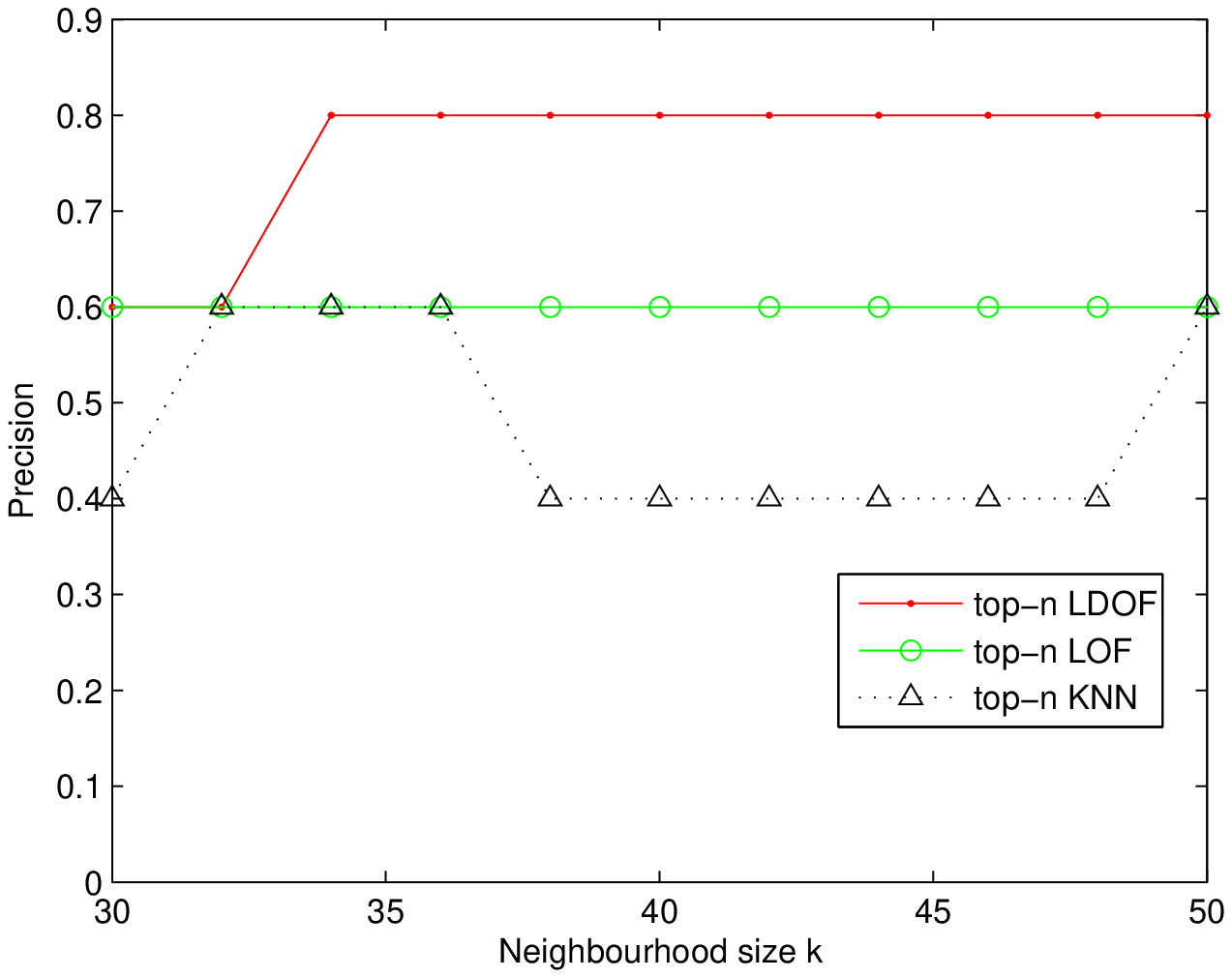,width=0.45\linewidth,height=0.35\linewidth}}

 \caption{Detecting precisions of top-$n$ LDOF,
top-$n$ KNN and top-$n$ LOF on (a) Synthetical
dataset, (b) WDBC dataset.} \label{fig:Exp12}
\end{figure*}

\paradot{Synthetic Data}
In Figure~\ref{fig:Demon}(b), there are 150 objects in cluster
$C_1$, 50 objects in cluster $C_2$, 10 objects in cluster $C_3$, and
4 additional objects $\{o_1, o_2, o_3, o_4 \}$ which are genuine
outliers. We ran the three outlier detection methods over a large
range of $k$. We use detection precision\footnotemark[2] to evaluate
the performance of each method. In this experiment, we set $n = 4$
(the number of real outliers). The experimental result is shown in
Figure~\ref{fig:Exp12}(a). The precision of top-$n$ KNN
becomes 0 when the $k$ is larger than 10 due to the effect of the
{mini-cluster} $C_3$ as we discussed in
Section~\ref{sec:problem}. For the same reason, the precision of
top-$n$ LOF dramatically descends when $k$ is larger than 11.
When the $k$ reaches 13, top-$n$ LOF misses all genuine
outliers in the top-4 ranking (they even drop out of {top-10}).
On the contrary, our method is not suffering from the effect of the
{mini-cluster}. As shown in the Figure~\ref{fig:Exp12}(a), the
precision of our approach keeps stable at $100\%$ accuracy over a
large neighbourhood size range (i.e. 20-50).

\paradot{Medical Diagnosis Data}
In real-world data repositories, it is hard to find a dataset
for evaluating outlier detection algorithms, because only for very
few real-world datasets it is exactly known which objects are
really behaving differently~\cite{Kriegel:08}. In
this experiment, we use a medical dataset, WDBC
(Diagnosis)\footnotemark[1], which has been used for nuclear feature
extraction for breast tumor diagnosis. The dataset contains 569
medical diagnosis records (objects), each with 32 attributes (ID,
diagnosis, 30 {real-valued} input features). The diagnosis is
binary: `Benign' and `Malignant'. We regard the objects labeled
`Benign' as normal data. In the experiment we use all 357 `Benign'
diagnosis records as normal objects and add a certain number of
`Malignant' diagnosis records into normal objects as outliers.
Figure~\ref{fig:Exp12}(b) shows the experimental result for adding
the first 10 `Malignant' records from the original dataset. Based on
the rule for selecting neighbourhood size, $k$, suggested in
Section~\ref{sec:property}, we set $k\geq 30$ in regards to the
data dimension. We measure the percentage of real outliers detected
in {top-10} potential outliers as detection
precision\footnotemark[2]. In the experiments, we progressively
increase the value of $k$ and calculate the detection precision for
each method. As shown in Figure~\ref{fig:Exp12}(b), the precision of
our method begins to ascend at $k = 32$, and keeps stable when $k$
is greater than 34 with detection accuracy of $80 \%$. In
comparison, the precision of the other two techniques are towed over
the whole $k$ value range.

To further validate our approach, we repeat the experiment 5 times
with a different number of outliers (randomly extracted from
`Malignant' objects). Each time, we perform 30 independent runs,
and calculate the average detection precision and standard deviation
over the $k$ range from 30 to 50. The experimental results are
listed in Table~\ref{tab:t1}. The bold numbers indicate that the
detection precision vector over the range of $k$ is statistically
significantly improved compared to the other two methods (paired
T-test at the 0.1 level).

\begin{table}[t]
\caption{The detecting precision for each method based on 30
independent runs.}\label{tab:t1} \centering
\begin{tabular}{|c|c|c|c|}
\hline Number of outliers &\multicolumn{3}{c|}{Precision (mean $\pm$ std.)} \\ \cline{2-4}
                &LDOF  &LOF    &KNN\\ \hline
1 & {\bf 0.29}$\pm$0.077 & 0.12$\pm$0.061 & 0.05$\pm$0.042 \\
2 & {\bf 0.33}$\pm$0.040 & 0.13$\pm$0.028 & 0.11$\pm$0.037 \\
3 & {\bf 0.31}$\pm$0.033 & 0.22$\pm$0.051 & 0.22$\pm$0.040 \\
4 & 0.35$\pm$0.022       & 0.27$\pm$0.040 & 0.26$\pm$0.035 \\
5 & {\bf 0.38}$\pm$0.026 & 0.28$\pm$0.032 & 0.28$\pm$0.027 \\ \hline
\end{tabular}
\end{table}

\begin{figure}
\begin{minipage}{0.45\textwidth}
\epsfig{file=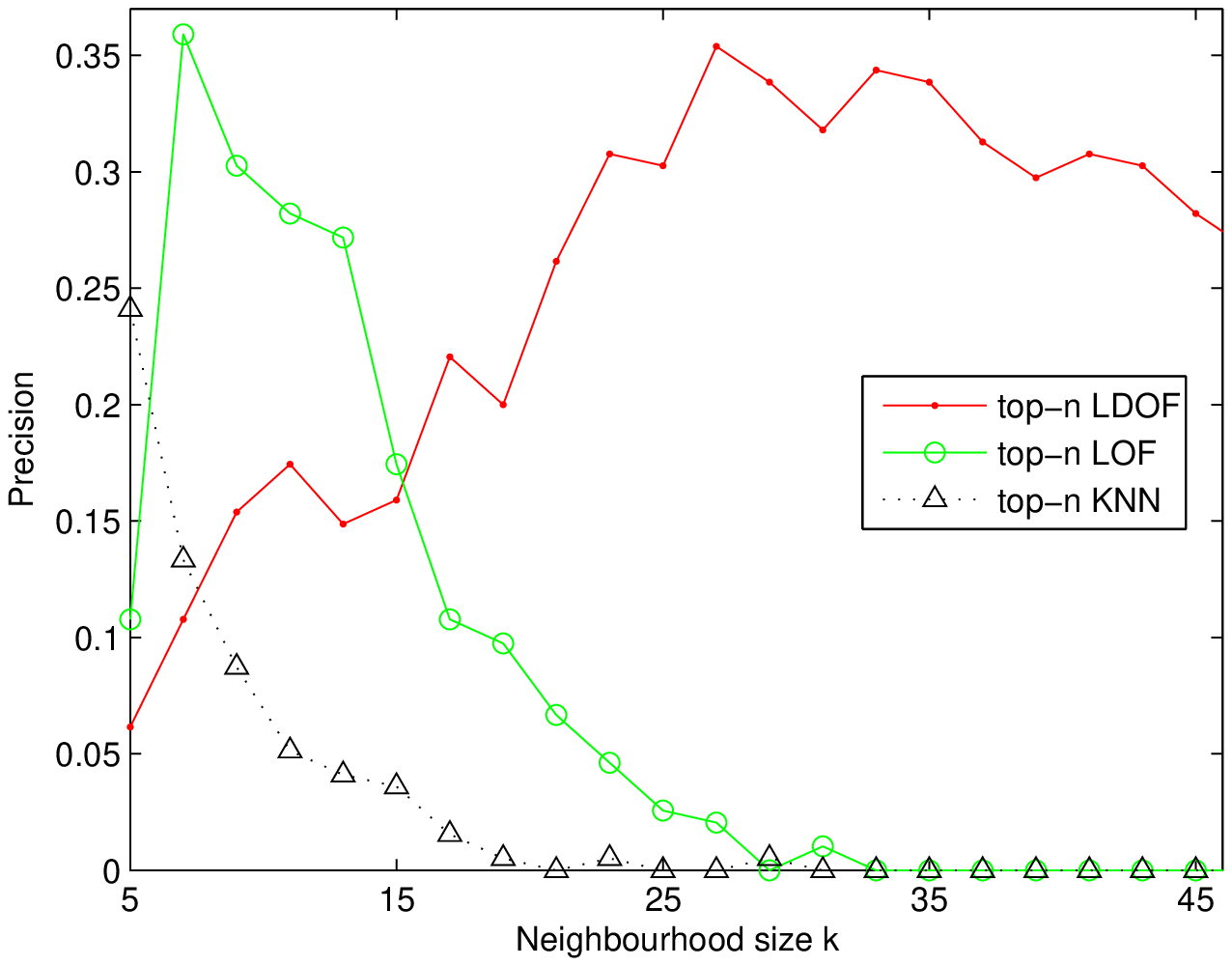,width=\linewidth}
\end{minipage}
\hfill
\begin{minipage}{0.5\textwidth}
\begin{tabular}{|c|c|c|}
\hline \multicolumn{3}{|c|}{Precision (mean $\pm$ std.)}\\ \hline
                LDOF  &LOF    &KNN                      \\ \hline
{\bf0.25}$\pm$0.081  &0.03$\pm$0.057    &0.08$\pm$0.114 \\ \hline
\end{tabular}
\caption{{\& Tab.\ref{tab:t2}.} Outlier detection precision over different
neighbourhood size for Shuttle dataset based on 15
independent runs.} \label{fig:Shuttle}\label{tab:t2}
\end{minipage}
\end{figure}

\paradot{Space Shuttle Data}
In this experiment, we use a dataset originally used for
classification, named Shuttle\footnotemark[3]. We use the testing
dataset which contains 14500 objects, and each object has 9
{real-valued} features and an integer label (1-7). We regard the
(only 13) objects with label 2 as outliers, and regard the rest of
the six classes as normal data. We run the experiment 15 times and
each time we randomly pick a sample of normal objects (i.e. 1,000
objects) to mix with the 13 outliers. The mean values of detection
precision of the three methods are presented in
Figure~\ref{fig:Shuttle}. As illustrated in
Figure~\ref{fig:Shuttle}, top-$n$ KNN has the worst performance
(rapidly drops to 0). {Top-$n$} LOF is better, which has a narrow
precision peak ($k$ from 5 to 15), and then declines dramatically.
{Top-$n$} LDOF has the best performance, as it ascends steadily and
keeps a relative high precision over the $k$ range from 25 to 45.
Table~\ref{tab:t2} shows the average precisions for the three
methods over 15 runs. The bold numbers indicate that the precision
vector is statistically significantly improved compared to the other
two methods (paired {T-test} at the 0.1 level).

\footnotetext[3]{The Shuttle dataset can also be downloaded from UCI
ML Repository.}

\section{Conclusion}\label{sec:conclusion}

In this paper, we have proposed a new outlier detection definition,
LDOF. Our definition uses a local distance-based outlier factor to
measure the degree to which an object deviates from its scattered
neighbourhood. We have analysed the properties of LDOF, including
its lower bound and false-detection probability. Furthermore,
a method for selecting $k$ has been suggested. In order to ease the
parameter setting in real-world applications, the
top-$n$ technique has been used in this approach.
Experimental results have demonstrated the ability of our new
approach to better discover outliers with high precision, and to
remain stable over a large range of neighbourhood sizes, compared to
top-$n$ KNN and top-$n$ LOF. As future work, we are
looking to extend the proposed approach to further enhance the
outlier detection accuracy for scattered real-world datasets.


\begin{small}

\end{small}

\end{document}